\documentclass[10pt,twocolumn,letterpaper]{article}
\usepackage[pagenumbers]{cvpr}

\usepackage{amsmath,amssymb,bm}
\usepackage{booktabs,multirow}
\usepackage{graphicx}
\usepackage{xcolor}
\usepackage{url}
\usepackage{tikz}
\usetikzlibrary{arrows.meta,positioning,calc,fit}
\usepackage[pagebackref=false,breaklinks,colorlinks,allcolors=cvprblue]{hyperref}
\definecolor{cvprblue}{rgb}{0.21,0.49,0.74}
\definecolor{ember}{RGB}{232,93,42}
\definecolor{tealc}{RGB}{15,139,141}
\definecolor{frz}{RGB}{54,88,160}

\newcommand{\ind}{\mathbb{1}}
\newcommand{\R}{\mathbb{R}}
\newcommand{\clip}{\operatorname{clip}}
\DeclareRobustCommand{\frozen}{\tikz[baseline=-0.55ex]{\foreach \a in {0,60,120} \draw[frz, line width=0.5pt] (\a:0.085) -- (\a+180:0.085); \foreach \a in {0,60,...,300} \draw[frz, line width=0.4pt] (\a:0.085) -- ++(\a+35:0.032) (\a:0.085) -- ++(\a-35:0.032);}}

\def\paperID{}
\def\confName{ECCV}
\def\confYear{2026}

\title{2nd Place Solution to the HANDS 2026 Workshop Challenge-Dexterous Grasp Motion Track:\\
Single-Shot Trajectory Warping for Grasp Motion Generation}

\author{Muneeb A. Khan$^{1}$ \qquad Woojin Kim$^{1}$ \qquad Shinwoo Kim$^{1}$\\Muhammad Munsif$^{1}$ \qquad Binod Bhattarai$^{2,3}$ \qquad Seungryul Baek$^{1}$\\
$^{1}$UNIST, South Korea \qquad $^{2}$University of Aberdeen, UK \qquad $^{3}$Fogsphere (Redev AI Ltd, UK)
}

\begin{document}

\maketitle

\begin{abstract}
This report describes our 2nd place solution to the HANDS 2026 workshop challenge (Dexterous Grasp Motion track) in conjunction with ECCV 2026. In this challenge, we address grasp motion generation for the 12-DoF LinkerHand O6, aiming to produce physically plausible reach-and-lift trajectories for unseen objects from randomized initial hand poses in simulation. This task is particularly challenging because each grasp requires a per-step policy to make approximately $70$ twelve-dimensional decisions, with errors accumulating over time, while test objects and physical dynamics may differ from those encountered during training. To address these challenges, we propose editing a single successful GraspM3 demonstration instead of generating the motion step by step: a policy observes the object once and outputs a 12-D warp of the demonstration, which is then replayed open-loop. Moreover, we train the warp policy with one-step PPO over all $4{,}824$ training objects in parallel. As a result, our method achieved success rates of $94.61\%$ on the easy track, the highest of all submissions, and $57.18\%$ on the hard track of the private test set.
\end{abstract}

\vspace{-0.5cm}

\section{Introduction}
\label{sec:intro}
Dexterous grasping is a prerequisite for general-purpose manipulation. A multi-fingered hand covers a far wider range of objects and grasp types
than a parallel-jaw gripper, but the added dexterity brings a high-dimensional action space and contact-rich dynamics that are hard to model~\cite{li2024tpgp}. Two developments have made data-driven approaches practical, including large synthetic corpora ~\cite{objaverse} that pair objects with hand poses or complete grasp motions, and massively parallel simulators~\cite{makoviychuk2021isaac}
that amortize exploration over thousands of objects at once. Combining them is nontrivial, because a corpus prescribes what the hand should do for one object instance, whereas a deployed policy must decide what to do for an unseen object from an arbitrary starting pose. The Dexterous Grasp Motion track~\cite{dexgraspmotion2026} is built around this gap. It builds on GraspM3~\cite{li2024tpgp,ye2025contact2motion}, retargeted from the Shadow Hand to the LinkerHand O6 consist of six actuated finger joints on a
free-floating wrist, so that every control step is a 12-D command (wrist
translation, wrist rotation, six finger targets). Training data covers $5{,}048$ ShapeNet~\cite{chang2015shapenet} objects with ${\sim}110$k trajectories; replaying them in Isaac Gym~\cite{makoviychuk2021isaac} and keeping the successful ones leaves $4{,}824$ objects and $52{,}718$ trajectories. Episodes start $15$--$20$\,cm from a randomly oriented object and succeed on a
$30$\,cm lift. Submissions are scored on private Objaverse~\cite{objaverse} objects on an \emph{easy} track (mass $20$\,g, friction $20$, as in training) and a \emph{hard} track (mass $\sim\mathcal{U}[20,200]$\,g, friction $\sim\mathcal{U}[1,20]$). The official baseline is behavior cloning on DexRep~\cite{liu2023dexrepnet} features ($23.93\%$).

Per-step policies face significant hurdles in this context, requiring approximately 70 twelve-dimensional decisions per grasp from a $1{,}757$-D observation, with compounding error. DemoGrasp~\cite{yuan2025demograsp} showed that neither is strictly necessary for the geometric phase of grasping: one successful demonstration, edited by a policy that decides \emph{where} the hand goes and \emph{how far} it closes, generalizes across objects when the editor is trained as a single-step MDP. We port this formulation to the O6 task and GraspM3 data. Our contributions are (i) an object-centric warp of one demonstration with position-only alignment, verbatim lift preservation, and proportional finger rescaling (\cref{sec:warp}); (ii) one-step PPO over all $4{,}824$ training objects in parallel with trajectory rotation (\cref{sec:train}); and (iii) an analysis of the easy-to-hard gap that an open-loop warp cannot close (\cref{sec:exp}).

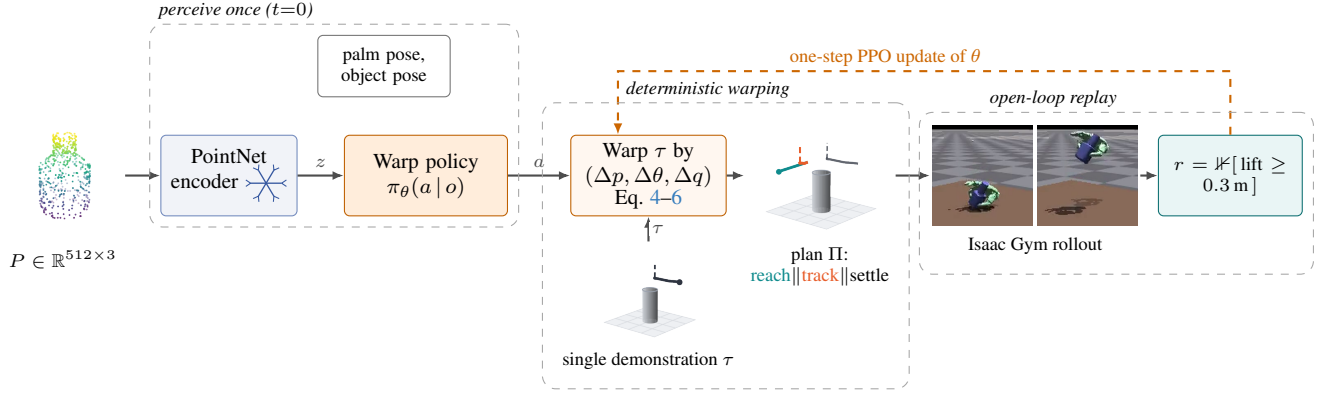
\begin{figure*}
    \centering
   \begin{tikzpicture}[x=0.95in, y=0.92in, font=\footnotesize,
  >={Latex[length=1.5mm, width=1.2mm]},
  blk/.style={draw=black!55, rounded corners=2pt, minimum height=0.42in, align=center, inner sep=2.5pt, fill=white, font=\footnotesize},
  frozen/.style={blk, fill=frz!7, draw=frz!70},
  learn/.style={blk, fill=orange!14, draw=orange!75!black},
  sim/.style={blk, fill=tealc!9, draw=tealc!80!black},
  grp/.style={draw=black!35, rounded corners=5pt, inner sep=4pt, dashed},
  glab/.style={font=\scriptsize\itshape, inner sep=1.5pt, anchor=south west},
  lab/.style={font=\scriptsize, fill=white, inner sep=1pt},
  sub/.style={font=\scriptsize, align=center},
  arr/.style={->, thick, black!70},
  thumb/.style={inner sep=0, outer sep=1pt}]
\node[thumb] (pc) at (0.34, 0) {\includegraphics[height=0.62in]{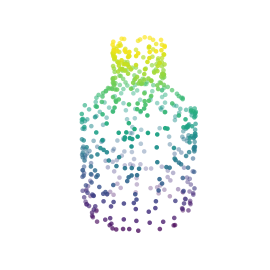}};
\node[sub, below=1pt of pc] (pclab) {$P\in\mathbb{R}^{512\times 3}$};
\node[blk, text width=0.62in, minimum height=0.32in, font=\scriptsize] (pose) at (2.1, 0.62) {palm pose,\\object pose};
\node[frozen, text width=0.64in] (enc) at (1.25, 0) {PointNet\\encoder~\frozen};
\node[learn, text width=0.78in] (pol) at (2.33, 0) {Warp policy\\$\pi_\theta(a\,|\,o)$};
% Changed: Shifted x-coordinate from 3.38 to 3.55
\node[learn, text width=0.74in, fill=orange!6, font=\scriptsize] (warp) at (3.55, 0) {\footnotesize Warp $\tau$ by\\$(\Delta p,\Delta\theta,\Delta q)$\\Eq.~\ref{eq:wrist}--\ref{eq:finger}};
% Changed: Shifted x-coordinate from 3.38 to 3.55
\node[thumb] (demo) at (3.55, -0.66) {\includegraphics[height=0.5in]{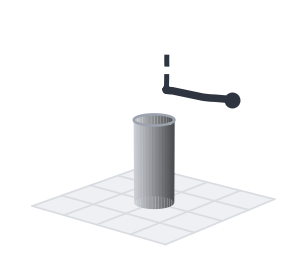}};
\node[sub, below=0pt of demo] (demolab) {single demonstration $\tau$};
% Changed: Shifted x-coordinate from 4.32 to 4.49
\node[thumb] (plan) at (4.49, 0) {\includegraphics[height=0.62in]{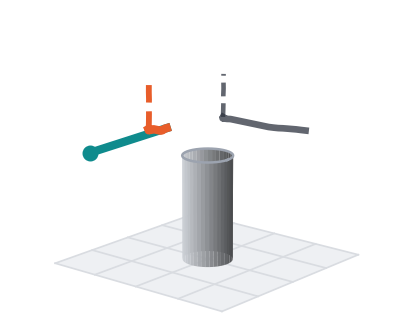}};
\node[sub, below=0pt of plan, align=center] (planlab) {plan $\Pi$:\\ \textcolor{tealc}{reach}$\|$\textcolor{ember}{track}$\|$settle};
\node[thumb] (sim) at (5.67, 0) {\includegraphics[height=0.52in]{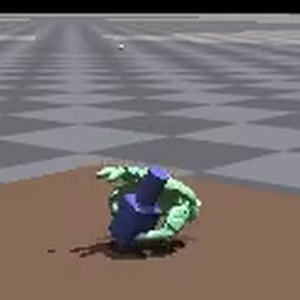}\hspace{2pt}\includegraphics[height=0.52in]{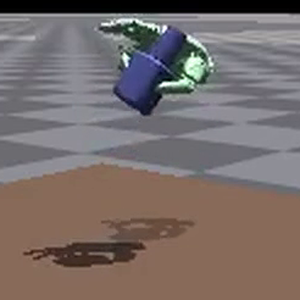}};
\node[sub, below=0pt of sim] (simlab) {Isaac Gym rollout};
\node[sim, text width=0.68in, font=\scriptsize] (rew) at (6.74, 0) {$r=\ind[\,\text{lift}\ge 0.3\,\text{m}\,]$};
\draw[arr] (pc) -- (enc);
\draw[arr] (enc) -- node[lab, above=2pt] {$z$} (pol);
\draw[arr] (pol) -- node[lab, above=2pt] {$a$} (warp);
\draw[arr] (demo) -- node[lab, right] {$\tau$} (warp);
\draw[arr] (warp) -- (plan);
\draw[arr] (plan) -- (sim);
\draw[arr] (sim) -- (rew);
% Changed: Increased xshift from 0.8in to 1.0in so the arrow still lands on the shifted warp box
\draw[arr, dashed, orange!80!black] (rew.north) -- ++(0, 0.36) -| ([xshift=1.0in]pol.north);
\node[lab, text=orange!80!black, anchor=south] at (4.85, 0.6) {one-step PPO update of $\theta$};
\node[grp, fit=(enc)(pol)(pose)] (g1) {};
\node[glab] at ([xshift=2pt]g1.north west) {perceive once ($t{=}0$)};
\node[grp, fit=(warp)(demo)(demolab)(plan)(planlab)] (g2) {};
\node[glab] at ([xshift=30pt]g2.north west) {deterministic warping};
\node[grp, fit=(sim)(simlab)(rew)] (g3) {};
\node[glab] at ([xshift=25pt]g3.north west) {open-loop replay};
\end{tikzpicture}
\caption{Overview of the trajectory warping pipeline. A frozen (\frozen) PointNet embeds the object point cloud $P$; with the palm and object poses, and the embedding $z$ is passed to the warp policy $\pi_\theta$ once per episode. }
\label{fig:overview}
\end{figure*}

\vspace{-0.1cm}

\section{Method}
\label{sec:method}
\Cref{fig:overview} shows the pipeline. An episode provides an object $o$ (initial pose and point cloud) and a randomized initial hand pose. The policy $\pi_\theta$ observes the scene once and outputs $a\in[-1,1]^{12}$; $a$ warps a single demonstration $\tau$ into a full trajectory, which is replayed open-loop, and the episode returns $r=\ind[\text{lift}\ge 0.3\,\text{m}]$. Learning is the contextual bandit
\begin{equation}
\max_\theta\ \mathbb{E}_{o\sim\mathcal{D},\,a\sim\pi_\theta(\cdot|o)}\big[r(o,a)\big],
\label{eq:obj}
\end{equation}
with no credit assignment across time.

\subsection{Object-centric demonstration}
\label{sec:demo}
\begin{figure}[t]
\centering
\includegraphics[width=0.7\columnwidth]{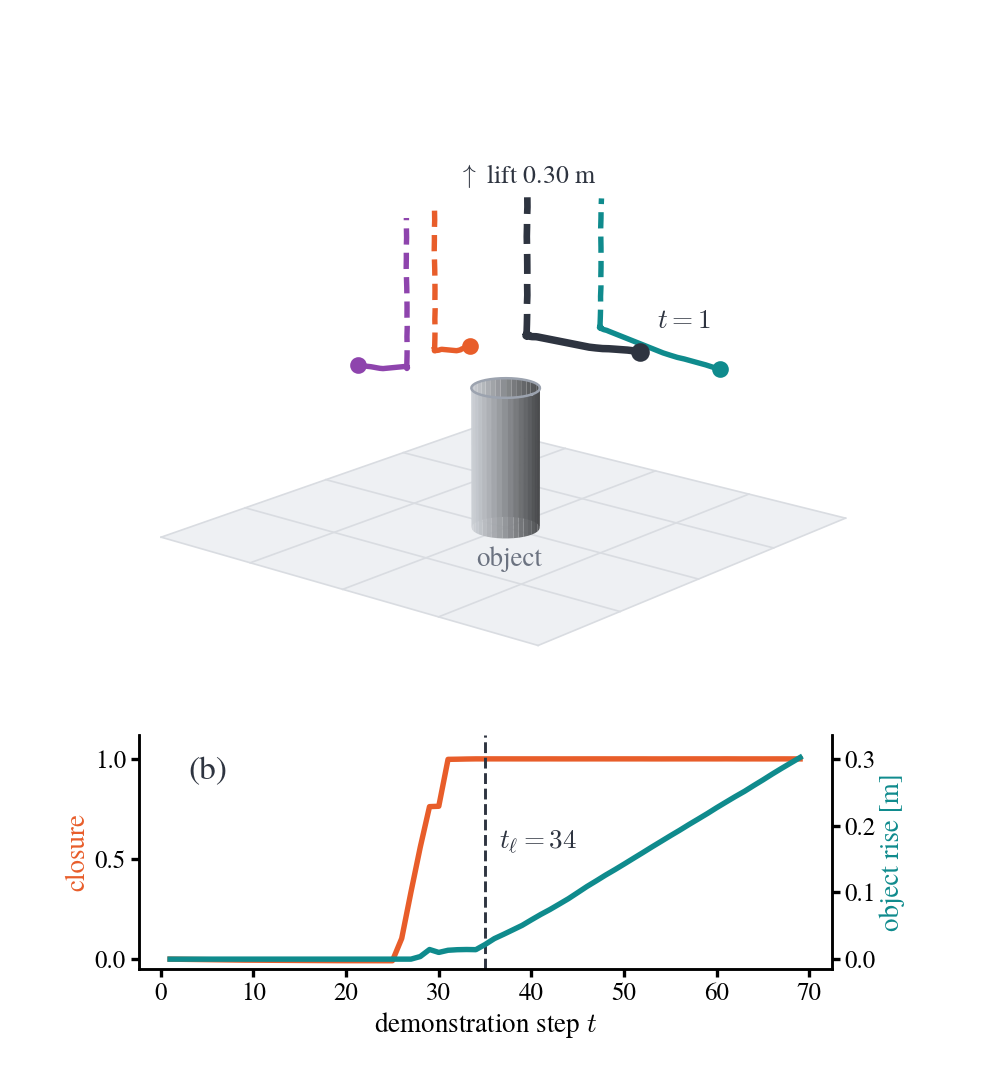}
\caption{A trajectory demonstration and sampled warps. (a) Palm path of $\tau$ (black; solid: approach and grasp, dashed: lift) and three warps (colors): $\Delta\theta$ rotates the approach about the object, $\Delta p$ shifts it, and the lift is copied verbatim (\cref{eq:lift}). (b) Replay in simulation: fingers close over steps $25$--$31$, lift onset $t_\ell=34$, $0.30$\,m rise by $t=69$.}
\label{fig:demo}
\end{figure}

We utilize a successful GraspM3 bottle grasp demonstration, verified via a single replay in Isaac Gym to ensure successful object lifting (\cref{fig:demo}). We record the \emph{commanded} actions; replaying the resulting physical states would re-introduce a one-step execution lag: $\tau=\{u_t\}_{t=1}^{T}$, $u_t=(x_t,r_t,q_t)\in\R^3\!\times\!\R^3\!\times\!\R^6$, $T=69$, with lift onset $t_\ell=34$ (first frame with object rise $>2$\,cm). The wrist command maps to the palm pose as $p_t=x_t+c$, $R_t=R_x(\phi_t)R_y(\theta_t)R_z(\psi_t)$ (intrinsic XYZ), with a constant offset $c$ measured at reset. We verified this convention numerically; Isaac Gym's extrinsic Euler helper is off by $5$\,cm and would corrupt every warped frame. The demonstration is expressed relative to its initial object \emph{position} only,
\begin{equation}
\tilde p_t = p_t - o_0^{\mathrm{demo}},\qquad \tilde R_t = R_t .
\label{eq:objcentric}
\end{equation}
Aligning to the object rotation as well drops no-warp replay success from $23.8\%$ to $4.8\%$: GraspM3 object rotations are near-random per trial ($\|\Delta R\|_F\in[0.77,2.83]$ within one object) and carry no approach information.

\subsection{Demonstration warping}
\label{sec:warp}
Following the DemoGrasp scales, the action is split into a wrist shift, a wrist rotation, and a finger offset,
\begin{equation}
\Delta p = s_p\,a_{1:3},\quad \Delta\theta = s_\theta\,a_{4:6},\quad \Delta q = s_q\,a_{7:12},
\label{eq:action}
\end{equation}
with $(s_p,s_\theta,s_q)=(0.05\,\text{m},\,1.57\,\text{rad},\,1.0\,\text{rad})$ and $R_\Delta=R_x(\Delta\theta_1)R_y(\Delta\theta_2)R_z(\Delta\theta_3)$.

\noindent\textbf{Wrist.} Pre-lift frames are rigidly warped in the object-centric frame, rotating the approach about the object and shifting it:
\begin{equation}
\hat p_t = R_\Delta\,\tilde p_t + \Delta p,\qquad \hat R_t = R_\Delta\,\tilde R_t\qquad (t<t_\ell).
\label{eq:wrist}
\end{equation}
Lift frames copy the demonstration's relative displacement onto the warped grasp pose, so the vertical motion that made the demonstration succeed is never rotated or scaled:
\begin{equation}
\hat p_t = \hat p_{t_\ell-1} + (\tilde p_t - \tilde p_{t_\ell-1})\qquad (t\ge t_\ell).
\label{eq:lift}
\end{equation}
The path returns to the world frame of the new episode (object at $o$) and to the action encoding: $p'_t=\hat p_t+o$, $x'_t=p'_t-c$, $r'_t=\mathrm{Euler}^{-1}_{XYZ}(\hat R_t)$.

\noindent\textbf{Fingers.} Let $q_1$ be the open pre-grasp pose and $q_{t_g}$, $t_g=t_\ell-1$, the closed grasp. The target grasp, $q^\star=\clip(q_{t_g}+\Delta q,\,q_{\min},\,q_{\max})$ is achieved by scaling the closing motion on a per-joint basis:
\begin{equation}
\lambda=(q^\star-q_1)\oslash(q_{t_g}-q_1),
q'_t=\begin{cases} q_1+\lambda\odot(q_t-q_1) & t<t_g\\ q^\star & t\ge t_g,\end{cases}
\label{eq:finger}
\end{equation}
so the closure morphs continuously into $q^\star$ with unchanged timing and the fingers keep squeezing through the lift.

\noindent\textbf{Trajectory synthesis.} The randomized reset pose $(x_0,R_0)$ is generally far from the warped first frame $(x'_1,R'_1)$. We prepend a reach segment (LERP on position, SLERP on orientation) of $n=\max\!\big(\lceil\|x'_1-x_0\|/0.01\rceil,\ \lceil\angle(R_0,R'_1)/0.1\rceil\big)$ steps, i.e.\ at most $1$\,cm and $0.1$\,rad per step, with fingers held at $q'_1$. The plan $\Pi=[\text{reach}_{1:T_r}\,\|\,\text{track}_{1:T}\,\|\,\text{settle}_{1:10}]$ is executed by sending each frame as an absolute 12-D target to the task's PD controller; the policy is never called again.

\subsection{Warp policy and one-step PPO}
\label{sec:policy}
The observation is the palm pose (7), the object pose (7), and a fixed-index subsample of $512$ points of the task's object-local cloud, encoded by the ShapeNet PointNet~\cite{qi2016pointnet} auto-encoder shipped with the DexRep baseline (frozen); only a linear projection to $z\in\R^{128}$ is trained. Actor and critic are separate MLPs $[1024,1024,512,512]$ with ELU and orthogonal initialization. The policy is a tanh-squashed Gaussian, $a=\tanh(\tilde a)$, $\tilde a\sim\mathcal{N}(\mu_\theta(o),\mathrm{diag}\,\sigma^2)$, with $\sigma$ learned (initialized at $0.8$) and the tanh Jacobian correction in $\log\pi_\theta$. With horizon one there is no bootstrapping: the return is $r$, $V_\phi(o)$ is a variance-reducing baseline, and $\hat A$ is $r-V_\phi(o)$ standardized over the batch. We minimize the clipped PPO~\cite{schulman2017ppo} objective
\begin{align}
\mathcal{L}=\;&\mathbb{E}\!\left[\max\!\big(-\hat A\rho_\theta,\,-\hat A\,\clip(\rho_\theta,1{-}\varepsilon,1{+}\varepsilon)\big)\right]\nonumber\\
&+c_v\,\mathbb{E}\!\left[(r-V_\phi(o))^2\right],
\label{eq:ppo}
\end{align}

where $\rho_\theta = \pi_\theta(a\vert{}o)/\pi_{\theta_{\mathrm{old}}}(a\vert{}o)$ denotes the policy ratio. We use a clipping threshold of $\varepsilon = 0.2$ and a value coefficient of $c_v = 2$. To ensure training stability, we clip gradients to a maximum norm of $1.0$. Furthermore, we employ an adaptive learning rate strategy based on the Kullback-Leibler (KL) divergence. The learning rate is scaled by a factor of $1.5$ to maintain a target KL divergence of $0.016$, and is strictly bounded within the interval $[10^{-5}, 10^{-2}]$.

\subsection{Large-scale training}
\label{sec:train}
Each iteration runs all $4{,}824$ training objects in parallel, one environment per object, sharded over three GPUs ($11.5$\,s per iteration); gradients, the KL statistic, and the advantage statistics are all-reduced, so the run equals a single $4{,}824$-environment batch. Because object scale is fixed at simulation build time, we rebuild the simulation every $K$ iterations with the next trajectory index, covering $51{,}659$ of $52{,}718$ trajectories ($98\%$). The training process converged over $4{,}000$ iterations (approximately 16 hours). The initial $2{,}000$ iterations utilized trajectory indices $0$--$3$, rotated every $500$ steps at an initial learning rate of $3\times10^{-4}$, then $2{,}000$ with indices $0$--$19$ rotated every $100$. Resuming the converged stage-1 policy at the default learning rate collapsed it from $80.3\%$ to $45.6\%$ training success within $24$ iterations; restarting at the rate the KL controller had settled on ($10^{-5}$) raised held-out success monotonically from $86.81\%$ to $90.98\%$.

\section{Experiments}
\label{sec:exp}
% \noindent\textbf{Setup.} Local evaluations are deterministic (mean action, one rollout per object) and run in chunks of $1{,}500$ objects per simulation, so peak GPU memory is independent of the object count ($\approx10.4$\,MiB per environment $+\,2.3$\,GB); leaderboard numbers are produced by the organizers on the private Objaverse set. Per the rules we disclose four configuration changes, all made by our entrypoint rather than by editing the task: the environment mode applies the full 12-D absolute target at every step; the per-step DexRep observation is replaced by the lightweight pose observation, since the policy is queried once; vision input is enabled; and the start-position canonicalization is disabled, a deterministic transform that the baseline path never triggers either. Initial poses and the success test are part of the task definition. We used only GraspM3 objects for training; the only pretrained component is the frozen PointNet already used by the official baseline.
\noindent\textbf{Setup.} We use three RTX~3090 GPUs. Local evaluation is deterministic (mean action, one rollout per object) in chunks of $1{,}500$
objects, so peak memory is independent of object count; leaderboard numbers come from the organizers. Per the rules, our entrypoint changes four settings
without editing the task: full 12-D absolute targets every step, the pose observation in place of per-step DexRep, vision enabled, and start-position
canonicalization disabled. Initial poses and the success test are unchanged. Training uses only GraspM3 objects; the sole pretrained component is the baseline's frozen PointNet.

\begin{table}[t]
\centering\footnotesize
\setlength{\tabcolsep}{5pt}
\begin{tabular}{lccc}
\toprule
Team (submission) & Easy [\%] & Hard [\%] & Rank \\
\midrule
HandsDown (2) & 91.29 & \textbf{77.45} & 1 \\
\textbf{UVLL HandDex (ours)} & \textbf{94.61} & 57.18 & 2 \\
J-Team & 87.20 & 43.88 & 3 \\
Perception & 68.80 & 23.98 & 4 \\
OverGrasp & 34.37 & 8.08 & 5 \\
HCBHand & 32.23 & 8.77 & 6 \\
\midrule
BC baseline (DexRep-TCN)~\cite{liu2023dexrepnet} & 23.93$^\dagger$ & -- & -- \\
\bottomrule
\end{tabular}
\caption{\textbf{Leaderboard} (private Objaverse test set, as of 1 Sept 2026): each team's highest-ranked submission, ranked by team. $^\dagger$Reported by the organizers on their demo split.}
\label{tab:leaderboard}
\end{table}

\begin{table}[t]
\centering\footnotesize
\setlength{\tabcolsep}{4pt}
\begin{tabular}{llcc}
\toprule
Checkpoint & Evaluation split & Objects & Success [\%] \\
\midrule
Final (stage 2) & held-out traj. (valid) & 4,502 & \textbf{90.98} \\
Final (stage 2) & training traj. (train) & 4,824 & 90.78 \\
Stage 1 & held-out traj. (valid) & 4,502 & 86.81 \\
64-object model & 64 held-out objects & 64 & 57.27$^\ddagger$ \\
\bottomrule
\end{tabular}
% \caption{\textbf{Local evaluation} (deterministic, one rollout per object). The valid split shares no grasp sequence with train, but $4{,}487$ of its $4{,}502$ objects also occur in train, so it measures generalization to unseen scale/start-pose combinations, not to unseen objects. $^\ddagger$Mean over 10 rounds on objects held out from that model's training; not comparable to the rows above.}
\caption{Local evaluation (deterministic, one rollout per object). The valid split shares no grasp sequence with train, but $4{,}487$ of its
$4{,}502$ objects appear in training. $^{\ddagger}$Mean over 10 rounds on held-out objects; not comparable to the rows above.}
\label{tab:local}
\end{table}

\noindent\textbf{Results.} \Cref{tab:leaderboard} compares teams. Our final submission scores $94.61\%$ on the easy track, the highest easy-track number of any submission, and $57.18\%$ on the hard track; HandsDown wins overall with $91.29/77.45$. Scaling the training set from $64$ objects (our second submission, $63.91/39.70$) to all $4{,}824$ added $+30.7$\,pp (easy) and $+17.5$\,pp (hard). \Cref{tab:local} reports the local evaluation: held-out trajectories score $0.2$\,pp \emph{higher} than training trajectories, i.e.\ there is no measurable overfitting to the trained trajectories, and stage 1 alone reaches $86.81\%$, so object coverage rather than iteration count is the main lever.

\begin{figure}[t]
\centering
\includegraphics[
    width=1\columnwidth,
    trim=2.9cm 0.6cm 0cm 0,
    clip
]{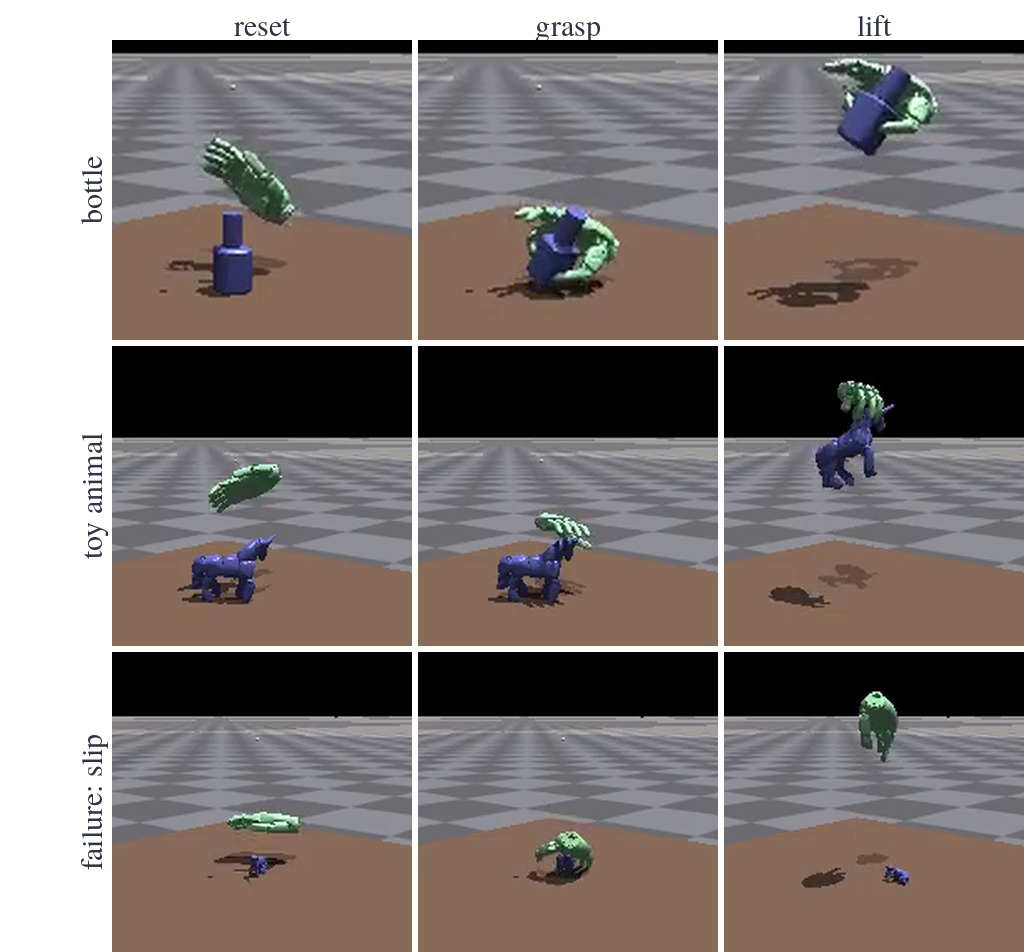}
\caption{Qualitative evaluation of the final policy in simulation. }
\label{fig:qual}
\end{figure}

\noindent\textbf{Qualitative results and ablations.} \Cref{fig:qual} shows rollouts on a bottle and a toy animal: the warp places the palm on the
surface, closes to a stable grasp, and the preserved lift completes the task. Failures follow one pattern: the object slips during the lift, and the plan proceeds without it. Three design choices had the largest effect: position-only alignment ($23.8\%$ vs.\ $4.8\%$ no-warp replay on $64$ objects), object coverage ($64\to4{,}824$ objects: $63.9/39.7\to94.6/57.2$), and the learning rate on resumption ($3\times10^{-4}$ collapses a converged policy; $10^{-5}$ improves it monotonically).

\noindent\textbf{Easy versus hard track.} We retain $60.4\%$ of our easy-track success on the hard track; HandsDown retains $84.8\%$. The gap is structural. (i) The plan is fixed before first contact, so an object that slides during the lift cannot be re-grasped. (ii) Mass and friction are invisible to the warp, which sees only geometry, and training used easy-track physics only. (iii) Grip force is inherited from the demonstration: finger targets are PD-tracked positions, and $\Delta q$ changes the shape of the grasp, not how hard it holds under ten times the load. The single decision therefore solves the geometric problem, \emph{where} and \emph{how} to grasp, while the remaining gap is dynamic and requires feedback.
\section{Conclusion}
\label{sec:conclusion}
We ported DemoGrasp's single-demonstration, one-step RL formulation to the LinkerHand O6 and scaled it to every GraspM3 training object. One demonstration and one 12-D decision per episode reach $94.61\%$ on unseen
Objaverse objects, the highest easy-track score in the challenge, though only $57.18\%$ under randomized mass and friction: the single decision solves the
geometric problem, while closing the dynamic gap requires feedback.
% In this report, we presented our solution to the Dexterous Grasp Motion track: a single-demonstration, one-step RL formulation of DemoGrasp ported to the LinkerHand O6 and scaled to every GraspM3 training object. One demonstration and one 12-D decision per episode reach $94.61\%$ on unseen Objaverse objects; our method ranked 2nd in the HANDS 2026 workshop challenge. %A primary limitation is its handling of complex dynamics: the open-loop warp retains only 60\% of its success rate when mass and friction parameters vary. Randomizing both inside the same one-step MDP and adding a closed-loop residual are the natural next steps.

% \section{Acknowledgement}
% [Funding / support acknowledgements, if any.]

{\small
\bibliographystyle{ieeenat_fullname}
\bibliography{refs}

@inproceedings{li2024tpgp,
  title={{TPGP}: Temporal-parametric optimization with deep grasp prior for dexterous motion planning},
  author={Li, Haoming and Ye, Qi and Huo, Yuchi and Liu, Qingtao and Jiang, Shijian and Zhou, Tao and Li, Xiang and Zhou, Yang and Chen, Jiming},
  booktitle={2024 IEEE International Conference on Robotics and Automation (ICRA)},
  pages={18106--18112},
  year={2024},
  organization={IEEE}
}

@article{ye2025contact2motion,
  title={{Contact2Motion}: Contact guided dexterous grasp motion generation with synergy embedded optimization},
  author={Ye, Qi and Li, Haoming and Liu, Qingtao and Jiang, Shijian and Zhou, Tao and Huo, Yuchi and Chen, Jiming},
  journal={The International Journal of Robotics Research},
  pages={02783649251364392},
  year={2025},
  publisher={SAGE Publications Sage UK: London, England}
}

@inproceedings{liu2023dexrepnet,
  title={{DexRepNet}: Learning dexterous robotic grasping network with geometric and spatial hand-object representations},
  author={Liu, Qingtao and Cui, Yu and Ye, Qi and Sun, Zhengnan and Li, Haoming and Li, Gaofeng and Shao, Lin and Chen, Jiming},
  booktitle={2023 IEEE/RSJ International Conference on Intelligent Robots and Systems (IROS)},
  pages={3153--3160},
  year={2023},
  organization={IEEE}
}

@article{yuan2025demograsp,
  title={{DemoGrasp}: Universal dexterous grasping from a single demonstration},
  author={Yuan, Haoqi and Huang, Ziye and Wang, Ye and Mao, Chuan and Xu, Chaoyi and Lu, Zongqing},
  journal={arXiv preprint arXiv:2509.22149},
  year={2025}
}

@article{makoviychuk2021isaac,
  title={{Isaac Gym}: High performance {GPU}-based physics simulation for robot learning},
  author={Viktor Makoviychuk and Lukasz Wawrzyniak and Yunrong Guo and Michelle Lu and Kier Storey and Miles Macklin and David Hoeller and Nikita Rudin and Arthur Allshire and Ankur Handa and Gavriel State},
  year={2021},
  journal={arXiv preprint arXiv:2108.10470}
}

@article{objaverse,
  title={{Objaverse}: A universe of annotated {3D} objects},
  author={Matt Deitke and Dustin Schwenk and Jordi Salvador and Luca Weihs and Oscar Michel and Eli VanderBilt and Ludwig Schmidt and Kiana Ehsani and Aniruddha Kembhavi and Ali Farhadi},
  journal={arXiv preprint arXiv:2212.08051},
  year={2022}
}

@article{qi2016pointnet,
  title={{PointNet}: Deep learning on point sets for {3D} classification and segmentation},
  author={Qi, Charles R and Su, Hao and Mo, Kaichun and Guibas, Leonidas J},
  journal={arXiv preprint arXiv:1612.00593},
  year={2016}
}

@article{schulman2017ppo,
  title={Proximal policy optimization algorithms},
  author={Schulman, John and Wolski, Filip and Dhariwal, Prafulla and Radford, Alec and Klimov, Oleg},
  journal={arXiv preprint arXiv:1707.06347},
  year={2017}
}

@article{chang2015shapenet,
  title={{ShapeNet}: An information-rich {3D} model repository},
  author={Chang, Angel X and Funkhouser, Thomas and Guibas, Leonidas and Hanrahan, Pat and Huang, Qixing and Li, Zimo and Savarese, Silvio and Savva, Manolis and Song, Shuran and Su, Hao and Xiao, Jianxiong and Yi, Li and Yu, Fisher},
  journal={arXiv preprint arXiv:1512.03012},
  year={2015}
}

@misc{dexgraspmotion2026,
  title={Dexterous Grasp Motion Challenge 2026 (HANDS Workshop at ECCV 2026)},
  author={{HANDS Workshop Organizers}},
  howpublished={\url{https://github.com/DexGraspMotionChallenge/DexGraspMotionChallenge2026}},
  year={2026}
}
}
\end{document}